\documentclass[conference]{IEEEtran}
\IEEEoverridecommandlockouts
\usepackage{cite}
\usepackage{amsmath,amssymb,amsfonts}
\usepackage{algorithmic}
\usepackage{graphicx}
\usepackage{textcomp}
\usepackage[caption=false]{subfig}
\usepackage{xcolor}
\usepackage{microtype}        
\usepackage{float}    
\usepackage{placeins} 
\usepackage{comment}
\usepackage{hyperref}
\usepackage{enumitem}
\usepackage{cite}
\usepackage{tabularx}
\usepackage{placeins}
\usepackage[compatibility=false]{caption}
\usepackage{booktabs}
\usepackage{float}

\usepackage[none]{hyphenat}
\begin{document}
\raggedbottom
\setlist[itemize,1]{leftmargin=\dimexpr 25pt \relax}   
\setlist[itemize,2]{leftmargin=\dimexpr 25pt \relax}   
\setlist[itemize,3]{leftmargin=\dimexpr 25pt \relax}   

\title{AI-Powered Image Analysis for Phishing Detection}

\author{
\IEEEauthorblockN{Kaushal Acharya$^{1}$ \quad Sunil Ale$^{1}$ \quad Rajan Kadel$^{2}$}

\author{\IEEEauthorblockN{Kaushal Acharya}
\IEEEauthorblockA{\textit{Melbourne Institute of Technology} \\
\textit{School of IT and Engineering}\\
Melbourne, Australia \\
MIT225353@stud.mit.edu.au}
\and
\IEEEauthorblockN{Sunil Ale}
\IEEEauthorblockA{\textit{Melbourne Institute of Technology} \\
\textit{School of IT and Engineering}\\
Melbourne, Australia \\
sale@academic.mit.edu.au}
\and
\IEEEauthorblockN{Rajan Kadel}
\IEEEauthorblockA{\textit{National Academy of Professional Studies} \\
\textit{School of IT}\\
Melbourne, Australia\\
rajan.kadel@naps.edu.au}
}}

\maketitle

\IEEEpubid{\begin{minipage}{\textwidth}\ \\ \\ \\ \\ \\ [10pt]
\centering
This is a preprint version of the paper accepted for the 9th International Conference on Inventive Computation Technologies (ICICT), 2026.
\end{minipage}}
\begin{abstract}
Phishing websites now rely heavily on visual imitation—copied logos, similar layouts, and matching colours—to avoid detection by text- and URL-based systems. This paper presents a deep learning approach that uses webpage screenshots for image-based phishing detection. Two vision models, ConvNeXt-Tiny and Vision Transformer (ViT-Base), were tested to see how well they handle visually deceptive phishing pages. The framework covers dataset creation, preprocessing, transfer learning with ImageNet weights, and evaluation using different decision thresholds. The results show that ConvNeXt-Tiny performs the best overall, achieving the highest F1-score at the optimised threshold and running more efficiently than ViT-Base. This highlights the strength of convolutional models for visual phishing detection and shows why threshold tuning is important for real-world deployment. As future work, the curated dataset used in this study will be released to support reproducibility and encourage further research in this area. Unlike many existing studies that primarily report accuracy, this work places greater emphasis on threshold-aware evaluation to better reflect real-world deployment conditions. By examining precision, recall, and F1-score across different decision thresholds, the study identifies operating points that balance detection performance and false-alarm control. In addition, the side-by-side comparison of ConvNeXt-Tiny and ViT-Base under the same experimental setup offers practical insights into how convolutional and transformer-based architectures differ in robustness and computational efficiency for visual phishing detection.
\end{abstract}


\section{Introduction} \label{sec:introduction}
One of the pervasive cybersecurity threats is phishing. It has exploited human trust through deceptive web interfaces and digital communications.  When studying traditional phishing detection systems, they have relied mainly on textual content or URL-based analysis. New age attackers now employ and use visual deception logos, colour schemes, and layout structures to mimic and present themselves as legitimate websites. They also evade text- and domain-based filters. These evolving threats have made image-based phishing detection an area of research, as it is emerging as a critical frontier in cybersecurity \cite{wangchuk2025multimodal,kritika2025comprehensive}.

The development and advancement of deep learning and computer vision in recent years have enabled automated extraction of spatial and contextual features. This is done by using webpage screenshots. Because of this, detection models are now capable of recognising and flagging subtle inconsistencies in visual composition. Convolutional Neural Networks (CNNs) and transformer-based vision architectures have shown and demonstrated the identification of such patterns. Most of the existing studies largely focus on accuracy but silently overlook and neglect other essential aspects, which are data set diversity, rigorous threshold optimisation, and model interpretability that are crucial for real-world deployment.

This paper addresses these areas not addressed by the existing literature. To achieve it, we developed an AI-powered image analysis framework, which is capable of systematically evaluating and analysing modern vision architectures for advanced phishing detection. For this, we have selected two representative vision models, ConvNeXt-Tiny \cite{liu2022convnet} and Vision Transformer (ViT-Base) \cite{dosovitskiy2020image}. We used these models to train and analyse their architectural strengths and feature extraction capabilities. Also, robustness against visually obfuscated phishing attacks. Selected models follow systematic processes in the order of dataset collection and construction, preprocessing, augmentation, model training, and threshold tuning and analysis for maximum output. Balancing recall and precision for deployment-relevant decision-making is the major result of threshold tuning, which is the objective of the paper. 

In summary, this study addresses the gap between the high reported accuracy in image-based phishing detection and the practical challenges of deploying such systems in real-world settings. While prior research has shown promising results, many studies rely primarily on accuracy and do not examine how models behave under different decision thresholds. The main purpose of this work is therefore to develop and evaluate a threshold-aware framework that provides a more realistic assessment of model performance. By systematically analysing precision, recall, and F1-score across varying thresholds and comparing ConvNeXt-Tiny and ViT-Base under the same experimental conditions, this study aims to offer clearer insight into both performance stability and practical applicability.

The contributions of this paper are as follows:
\begin{itemize}
    \item A threshold-aware evaluation methodology that optimises precision, recall, and F1-score for real-world deployment. An experimental analysis highlighting architectural trade-offs between accuracy and computational efficiency
    \item An end-to-end framework for image-based phishing detection using webpage screenshots, comparative evaluation of ConvNeXt-Tiny and transformer-based vision architectures;
\end{itemize}

The remainder of this paper is structured as follows. Section~\ref{sec:RW} reviews related works on phishing detection. Section~\ref{sec:motivation} defines the problem statement and outlines the contributions. Section~\ref{sec:methodology} describes the methodology used in this research. Section~\ref{sec:results} presents the experimental evaluation and comparative analyses. Finally, Section~\ref{sec:conclusions} concludes the paper with key findings and future directions.

\section{Related Works} \label{sec:RW}
This section reviews recent literature on phishing detection techniques across text, URL, and image modalities.
\subsection{Text-based Phishing Detection}

Traditional phishing detection using text has primarily relied on linguistic and semantic features that were extracted from email bodies and subjects. Natural language processing, combined with feature engineering, enables classifiers to distinguish between legitimate and phishing messages with high accuracy. Some of the common techniques used were tokenisation, Term Frequency-Inverse Document Frequency (TF-IDF) weighting, lemmatisation, Chi-square, principal component analysis, and latent semantic analysis to generate compact but discriminative feature sets \cite{gualberto2020answer}. Classifiers such as XGBoost, Random Forest, and Support Vector Machines (SVMs) have demonstrated strong performance on benchmark datasets when trained on engineered text features \cite{gualberto2020answer}. Latest reviews have highlighted that TF-IDF, n-grams, and word embeddings are still dominant techniques in text-based phishing detection, along with new CNNs and Recurrent Neural Networks (RNNs) gaining adoption recently \cite{salloum2022systematic}. Verma et al. further showed that preprocessing techniques such as stemming, stopword removal, and n-gram extraction improve detection, though false positives persist due to overlaps with legitimate marketing emails \cite{verma2020email}. Despite these advances, text-based models remain insufficient against modern attacks that embed logos, images, or obfuscated scripts, underscoring the need for multi-modal defences.

\begin{table*}[htbp]
\centering
\caption{Summary of Image-Based Phishing Detection Approaches and Datasets}
\label{tab:image_based_combined}
\begin{tabularx}{\textwidth}{p{3cm} p{1cm} p{8cm} p{4.5cm}}
\toprule
\textbf{Method / Model} & \textbf{Accuracy} & \textbf{Dataset (Name \& Size)} & \textbf{Key Observation(s)} \\
\midrule
CNN-based models~\cite{zieni2023phishing} & $>95\%$ & Balanced URL dataset (5,000 phishing + 5,000 legitimate) & Dataset dependency issues \\

CNN + LSTM hybrid~\cite{alshingiti2023deep} 
& $>97\%$ 
& Public phishing benchmark datasets (size not explicitly reported) 
& High recall; computationally expensive \\

Phishpedia (Faster-RCNN + Siamese)~\cite{lin2021phishpedia} 
& $99.2\%$ 
& Phishpedia benchmark ($\sim$30K phishing + $\sim$30K benign webpages) 
& Detected 1700+ zero-day sites; runtime of 0.19 s/page \\

CNN visual similarity~\cite{saeed2022visual} 
& $\sim96\%$ 
& Screenshot-based phishing datasets (e.g., Phish-IRIS $\sim$2,852 images) 
& Strong logo recognition; limited dataset diversity \\

Vision GNN~\cite{lindamulage2023vision} 
& $>97\%$ 
& Vision-based phishing benchmark datasets (dataset size not clearly specified) 
& Captures relational structure among webpage elements \\

Phish-IRIS (compact descriptors)~\cite{dalgic2018phish} 
& $\sim92\%$ 
& Phish-IRIS dataset (1,313 training + 1,539 testing images) 
& Lightweight approach; less accurate than CNNs \\
\bottomrule
\end{tabularx}
\end{table*}

\subsection{URL-based Phishing Detection}

URL-based phishing detection is one of the widely studied areas. It is studied using lexical, structural, and domain-related attributes. One of the recent studies applied a dataset of over 11,000 URLs with 33 extracted features, such as special characters, domain age, and protocol type. These are used to evaluate classifiers such as k-Nearest Neighbour, Logistic Regression, Decision Tree, Random Forest, Gradient Boosting, Naïve Bayes, and Support Vector Classifier. Some of the combinations achieved strong results, with Random Forest reaching 96.77\% accuracy and hybrid exceeding 98\% accuracy \cite{karim2023phishing}. Also, PhishDef demonstrated that carefully engineered lexical features alone can handle full feature sets, achieving $96$--$98\%$ accuracy. Algorithms such as AROW, while remaining resilient to noisy training data and obfuscation techniques, are thus suitable for real-time browser-side detection \cite{le2011phishdef}. Complementing these approaches, Sánchez-Paniagua et al. introduced the PILU-90K dataset of 90,000 login and phishing URLs, showing that many existing models trained on homepage URLs suffer high false-positive rates when confronted with real login pages. 
TF-IDF N-gram features achieved 96.5\% accuracy with logistic regression, while CNN models were able to demonstrate competitive performance \cite{sanchez2022phishing}. In total, these findings point towards that ensemble and lexical-feature-based methods provide strong accuracy, but there are certain challenges in addressing real-case login scenarios and tackling evolving obfuscation. This is actually a real motivating factor for the integration of adaptive and deep learning solutions.

\subsection{Image-based Phishing Detection}
Recent attackers target more in the form of visual attacks. Image-related techniques have gained more interest. Attackers are starting to incorporate the use of logos and graphics in fooling and tricking the user. The major reason for this is that it can easily evade text and URL-filtering mechanisms. Surveys in phishing-related detection emphasise the fact that computer vision has crossed the boundaries of being an essential part of modern technology systems, with the use of CNNs in phishing screenshot datasets being seen across the board with classification accuracies of over 95\% in phishing-related datasets \cite{zieni2023phishing,alshingiti2023deep}. They incorporate the detection of logos, designs, and colour patterns. The capture of the above is very helpful in the aspect of being very hard to interpret by the text-related classifiers in terms of character replacement and obfuscation by scripts. However, despite the improvements found in the models, they are seriously reliant on the dataset they are trained on. They perform poorly if the training datasets lack adequate information in terms of new brand names, or rather, creatively changed logos and banners. This dependency raises questions about scalability, as obtaining large, labelled image datasets that reflect the evolving phishing landscape remains a persistent challenge. These limitations indicate that although CNN-based detection has satisfactory baseline performance, it is not capable enough to address zero-day threats and rapidly evolving visual deception techniques and dynamic strategies \cite{saeed2022visual}.

Phishpedia, proposed by Lin et al., is a hybrid approach efficient in visually scanning, analysing, and detecting phishing websites through the comparison and cross-validation of logos and login form boxes in the screenshots of webpages \cite{lin2021phishpedia}. The proposed method combines the use of a Faster-Region-based CNN (RCNN) model for logo and UI component detection and the Siamese network for brand validation and cross-validation for the accurate differentiation between legitimate and phishing/suspect domains. Conventional techniques for phishing website detection necessitate the need for large amounts of labelled phishing data, while Phishpedia employed the utility of transfer learning through logos of brands, ensuring flexibility in detecting new targets without the need for further training on phishing data. Experimental analysis on more than 30,000 phishing and legitimate webpages demonstrated the effectiveness of Phishpedia in outperforming other conventional techniques such as EMD, PhishZoo, and LogoSENSE in identifying phishing webpages at a 99.2\% accuracy in an absolute runtime of  0.19 seconds per webpage. It was also capable of resisting malicious attacks. This clearly testified to the efficacy of Phishpedia in realistic applications. This also identified more than 1,700 phishing webpages in a short span of 30 days, astonishingly including some as zero-day attacks unidentified by other networks.

Recently, this area has moved beyond traditional CNNs and started looking at how webpages are structured. Lindamulage et al. used Vision GNNs to model relationships between visual elements, reaching above 97\% accuracy on benchmark datasets \cite{lindamulage2023vision}. Saeed also has proposed another visual similarity approach that uses CNN-based embeddings to detect brand duplication, impersonation, with results around 96\% \cite{saeed2022visual}. Earlier methods and models, such as  Phish-iris focused on lightweight and surficial visual descriptors for phishing detection, offering faster, easier, and simpler solutions, though with lower accuracy (about 92\%) compared to CNN-based models \cite{dalgic2018phish}. These studies demonstrated that image-based methods generally outperform traditional techniques; however, some issues remain. Dataset diversity, adversarial changes, and scalability are still challenges, so more research is needed.

Table~\ref{tab:image_based_combined} summarises the accuracy, datasets used and key observations of the various studies. This highlights the differences and performances across recent image-based phishing detection models and approaches. It is observed that individual study shows promising results, their datasets, efficiency and limitations vary.
  
\subsection{Metrics Observations}
Accuracy is still the most common metric in phishing detection research. Accuracy alone is not reliable for imbalanced datasets, where benign samples are much higher. Precision, recall, and F1-score give a better picture as they show how the model handles and analyses false positives and false negatives. When reviewed, many papers do not report these metrics in a consistent way. For example, Alshingiti et al. \cite{alshingiti2023deep} reported an F1-score of 0.96 for their CNN–LSTM model. Vision GNN studies showed high precision and recall, but did not always include F1-scores. Approaches such as  Phishpedia, primarily focused on accuracy and runtime.

From the above discussion, it is clear why precision, recall, and F1-score together are more helpful than a singular metric, suggesting the efficiency and capability of the model. The next section articulates the problems we are addressing in this paper.

\section{Problem Statement} \label{sec:motivation}

Most of the recent image-based phishing detection methods, like CNN hybrids, Phishpedia, and Vision GNN, have suggested strong accuracy but most of them are lacking decision-threshold analysis. These models might do well on balanced datasets whereas a higher chance of failure is possible in real cases where benign samples are much higher. Alshingiti et al.~\cite{alshingiti2023deep} reported F1-scores in their research, but many papers still do not show how performance varies at different thresholds mention of the decision threshold has made the results vague, harder to reproduce and harder to understand the progress of the models. The next issue that comes to the surface is that the use of accuracy is really heavy as the main metric. When there are imbalanced datasets, accuracy can be misleading and give more ambiguous information. The model can predict most samples as benign and still get high accuracy. Precision, recall, and F1-score are more meaningful, but are not reported clearly.

Image-based models like CNNs and GNNs have a heavy dependency on training data quality. If the dataset does not include recent brands or altered logos, these models struggle to generalise and detect properly. Tricks such as Zero-day, changing colours, layouts, or hiding elements to evade detection are now the favourite of attackers. All of these issues suggest a need for phishing detection systems that use image analysis and also include threshold-based evaluation. Models should report precision, recall, and F1-scores and also be able to handle class imbalance and adversarial changes. Addressing these gaps is key to building systems that work reliably in real-world conditions.

\subsection{Challenges in the Current State of Phishing Detection}

Websites now have visual imitation rather than simple text tricks, which is why phishing attacks are becoming harder to detect because many attackers copy logos, colour schemes, and layouts, which help them to bypass systems that only rely on analysing  URLs or text content. Some phishing pages have embedded text inside images, which makes it difficult for string-matching or DOM-based methods to identify potential content. Deep learning models can help, but they depend heavily on the quality and variety of the training data. Non-inclusion of newly emerging brands or altered logos can make these models fail to generalise to real-world phishing attempts.

Another major challenge is that many detection models and systems do not incorporate new attack patterns. Phishing kits and techniques are very dynamic. The mindset of attackers is changing every hour. To avoid detection, Attackers are using zero-day visual changes, such as modified layouts or redesigned icons. Without ongoing updates or feedback loops, model performance and detection capability drop over time. An imbalance between datasets is also a persistent issue, as there are more legitimate webpages outnumbering phishing ones. These are the reasons why accuracy is less reliable, and there is a need for metrics such as precision, recall, and F1-score. In sum, these challenges clearly suggest why phishing detection systems must be able to handle visual variation and changes, evolving attack styles, and imbalanced datasets to remain effective in real-world deployments.

\subsection{Our Contributions}
This research will provide significant contributions to image-based phishing detection. Firstly, it has provided an end-to-end framework that analyses, classifies phishing and legitimate pages using webpage screenshots. As the framework consists of different yet simple workflows, which consist of stages like data collection, preprocessing, model setup, training, and evaluation. As a single system, the processes are easy to reproduce and follow. Major focus includes a threshold-aware evaluation, which was a missing part in most of the recent models and research.

Secondly, the study compares two well-known vision models, ConvNeXt-Tiny and ViT-Base, under the same criteria and conditions. This parallel comparison is to fetch the strengths and weaknesses of each model, particularly how they learn and analyse visual patterns such as logos, layout structure, and colour schemes. The analysis highlights where convolutional models still perform better and where transformers may struggle without larger datasets.

Thirdly, the work uses transfer learning with ImageNet-pretrained weights. This streamlines the training, making it faster and handling limited phishing data more effectively. The results show that this approach improves stability and performance during training, particularly for the convolutional model.

Another major contribution is the use of threshold optimisation for the model, which tests different decision thresholds to find the best operating and performing point for real-world use. This step improves the balance between precision and recall and gives a clearer picture of how the models behave once deployed.

In summary, the study provides practical guidance and step-by-step explanations on building and evaluating image-based phishing detection systems. It aims to support future work by releasing the curated dataset so that other researchers can compare methods and explore new solutions.

\section{Methodology} \label{sec:methodology}
\begin{figure*}[htbp]
\centering
\includegraphics[width=0.95\textwidth]{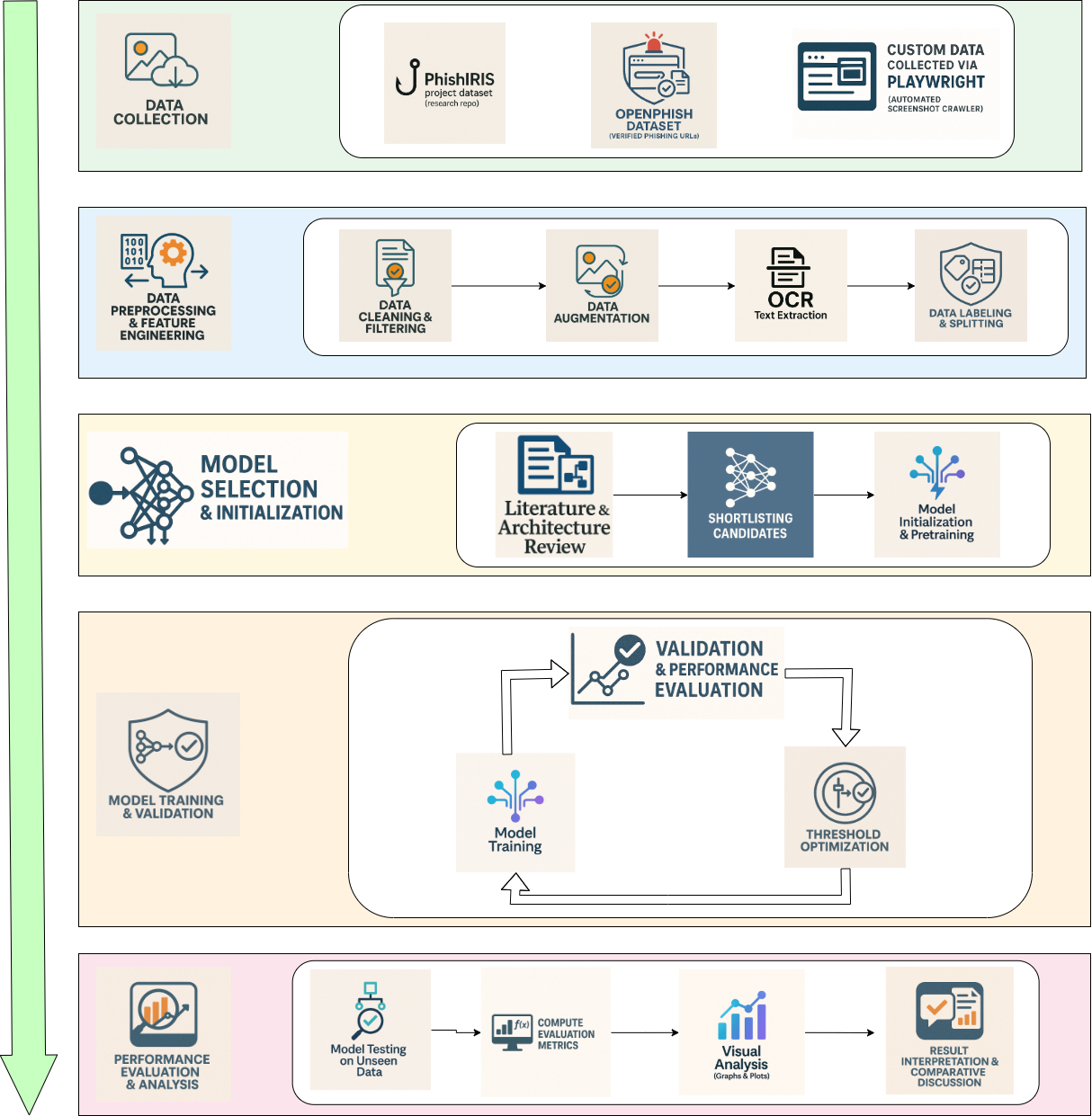}
\caption{Methodology applied during the study.}
\label{fig:methodology}
\end{figure*}

This section explains the steps involved in the study Fig.~\ref{fig:methodology} is the visual representation of the steps involved.

\subsection{Stage 1: Data Collection}
As data constitutes the core component of the system, it is always essential to maintain a clean and updated dataset. However, public datasets were limited in terms of scale and recency. To address this limitation, a Python-based data collection pipeline was developed using Playwright (a software testing and browser automation framework) to automatically capture Screenshots of webpages. For this, the latest URL feeds from OpenPhish were used as a source for the script\cite{OpenPhish2025Database}. This process resulted in 18,432 screenshots. In addition, 10,444 mixed screenshots were used from the PhishIRIS dataset \cite{Shahane2025PhishIRIS} to increase data diversity. In total, 28,876 labelled webpage screenshots were in the dataset, which gave a balanced and up-to-date dataset for training, validating and testing the system.

\subsection{Stage 2: Data Preprocessing and Feature Engineering}

In this stage, all the gathered screenshots were resized to 224*224 pixels to match the input size of ImageNet-based models. Using a hash algorithm, the dataset was cleaned by detecting duplicate and corrupted images, which were removed from the final datasets. To make the dataset even diverse, data augmentation steps like random horizontal flips, brightness changes (±15\%), Gaussian blur, and random cropping were performed. Finally, the dataset was split into 80\% training, 10\% validation, and 10\% testing, keeping both classes balanced. 

\subsection{Stage 3: Model Selection and Initialisation}
For this study, we have chosen two models: ConvNext-Tiny and Vision Transformer (ViT-Base) on their basis of performance, scalability, and deployment needs \cite{lin2021phishpedia}. ConvNext-Tiny is a modern CNN-based model that strongly focuses on visual recognition, while ViT-Base learns broader image patterns using self-attention. ConvNeXt-Tiny updates standard CNN ideas to reach competitive accuracy while staying efficient \cite{liu2022convnet}. Its local feature learning strength makes it suitable for detecting the visual details such as Brand-imitating logos, visually similar layouts, and embedded textual information, which are often used in phishing websites. 

On the other hand, ViT-based architectures can analyse the image patches while capturing long-range contextual dependencies, enabling effective modelling of global webpage structure \cite{dosovitskiy2020image}. This ability can be leveraged for phishing webpage detection, as global layout and structural cues help distinguish phishing pages from legitimate websites. Previous research indicates that transformer architectures can perform strongly in visual phishing detection, justifying the use of ViT-Base as a comparison model \cite{lin2021phishpedia}. Evaluating both models within the same experimental setup allows this study to examine the trade-offs between local feature extraction and global context modelling in detecting phishing webpages.

\subsection{Stage 4: Model Training and Validation}
Using labelled screenshots, models were trained under a supervised learning framework to classify phishing and legitimate. Binary cross-entropy was employed as the loss function to quantify classification error, while the Adam optimiser was used to update model parameters during training. Hyperparameters, including the learning rate, batch size and dropout, were tuned during the training phase. For the classification of these binary images, the following choice configurations follow the standard practice \cite{alshingiti2023deep}.

Validation was performed on each epoch while observing model overfitting and generalisation on unseen data. Metrics score, including precision, recall, and F1-score, were tracked and analysed to see how the model is working. A post-training threshold optimisation step evaluated multiple classification thresholds to achieve the best F1-score while reducing false positives, consistent with earlier deployment-oriented study \cite{lin2021phishpedia}.

\subsection{Stage 5: Performance Evaluation and Analysis}
The performance of model was evaluated using precision, recall, and F1-score, as these metrics provide a clearer picture than accuracy in phishing detection. Accuracy can be misleading under class imbalance, whereas precision and recall highlight false-positive and false-negative behaviour. The F1-score offer a balanced view of both aspects. This matches earlier phishing research that relies more on class-sensitive metrics than overall accuracy. Observing these metrics enabled a clear understanding of the model's response to phishing and legitimate websites \cite{powers2020evaluation}.

A threshold optimisation step has also been used to make the experiment more suitable for implementation. Rather than using a standard threshold, various values for the threshold were tested to observe the changes in the score. This made it easier to assess at which stage the classifier maintained a high level of recall and lowered the number of false positives. While the performance of ConvNeXt-Tiny was more stable under varied threshold levels, the ViT-Base model needed more optimisation. These aspects have provided a better understanding of which classifier is more controllable and how they will work in a real-world scenario \cite {saito2015precision}.

\section{Results and Analysis}\label{sec:results}
In this section, the outcome obtained by utilising the proposed methodology in Section~\ref{sec:methodology} will be explained. After the dataset is constructed and the preprocessing is done, ConvNeXt-Tiny and ViT-Base were trained by using the transfer learning method initialised by ImageNet. The trained models would then make use of the threshold-aware analysis. The outcome focuses on precision, recall, and F1-score.

\begin{figure}[htp]
\centering
\includegraphics[width=8.75cm, height=6cm]{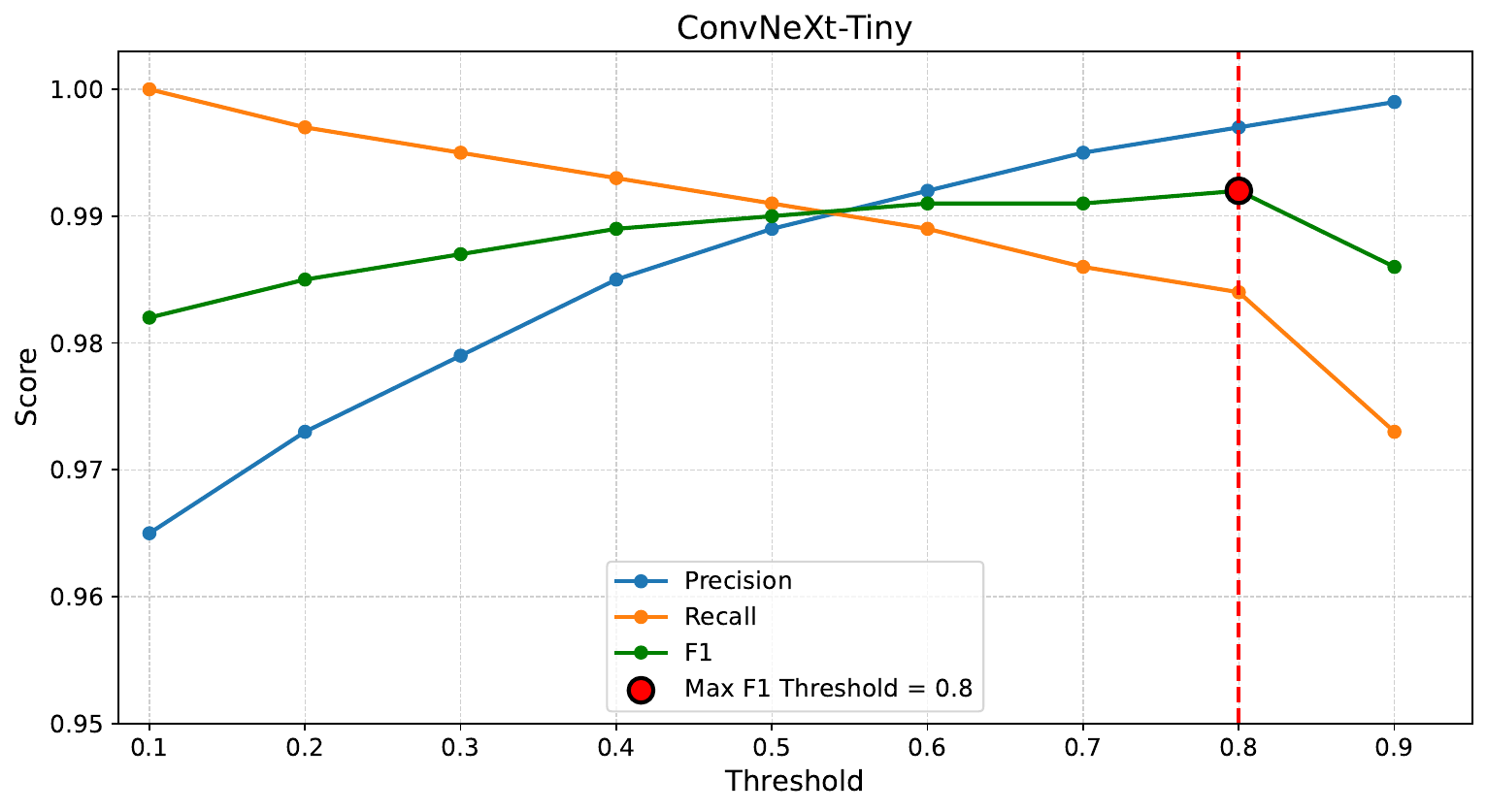}
\caption{Threshold analysis of ConvNeXt-Tiny.}
\label{fig:threshold-tiny}
\end{figure}

Table~\ref{tab:convnext_threshold} and Fig.~\ref{fig:threshold-tiny} illustrate precision, recall, and F1-score against the various threshold values for the \textit{ConvNeXt-Tiny} model. From the results, it can be seen that precision values increase when the threshold values increase. Therefore, the more accurate the model is in making predictions without having false positives. The value of the recall indicator decreases because the model becomes less vigilant about phishing sites at higher threshold values. The F1-score obtains its maximum value when the threshold value is 0.8. At this point, a balance between identifying phishing attempts and sending undue warnings can be obtained. The discovery of this threshold value becomes significant since it can make the model more accurate when it is put into practice.
\begin{table}[htp]
\centering
\caption{Threshold-based performance of ConvNeXt-Tiny}
\label{tab:convnext_threshold}
\begin{tabular}{c c c c}
\hline
\textbf{Threshold} & \textbf{Precision (\%)} & \textbf{Recall (\%)} & \textbf{F1-score (\%)} \\
\hline
0.1 & 0.965 & 1.000 & 0.982 \\
0.2 & 0.973 & 0.997 & 0.985 \\
0.3 & 0.979 & 0.995 & 0.987 \\
0.4 & 0.985 & 0.993 & 0.989 \\
0.5 & 0.989 & 0.991 & 0.990 \\
0.6 & 0.992 & 0.989 & 0.991 \\
0.7 & 0.995 & 0.986 & 0.991 \\
\textbf{0.8} & \textbf{0.997} & \textbf{0.984} & \textbf{0.992} \\
0.9 & 0.999 & 0.973 & 0.986 \\
\hline
\end{tabular}
\end{table}

\begin{figure}[htp]
\centering
\includegraphics[width=8.55cm, height=6cm]{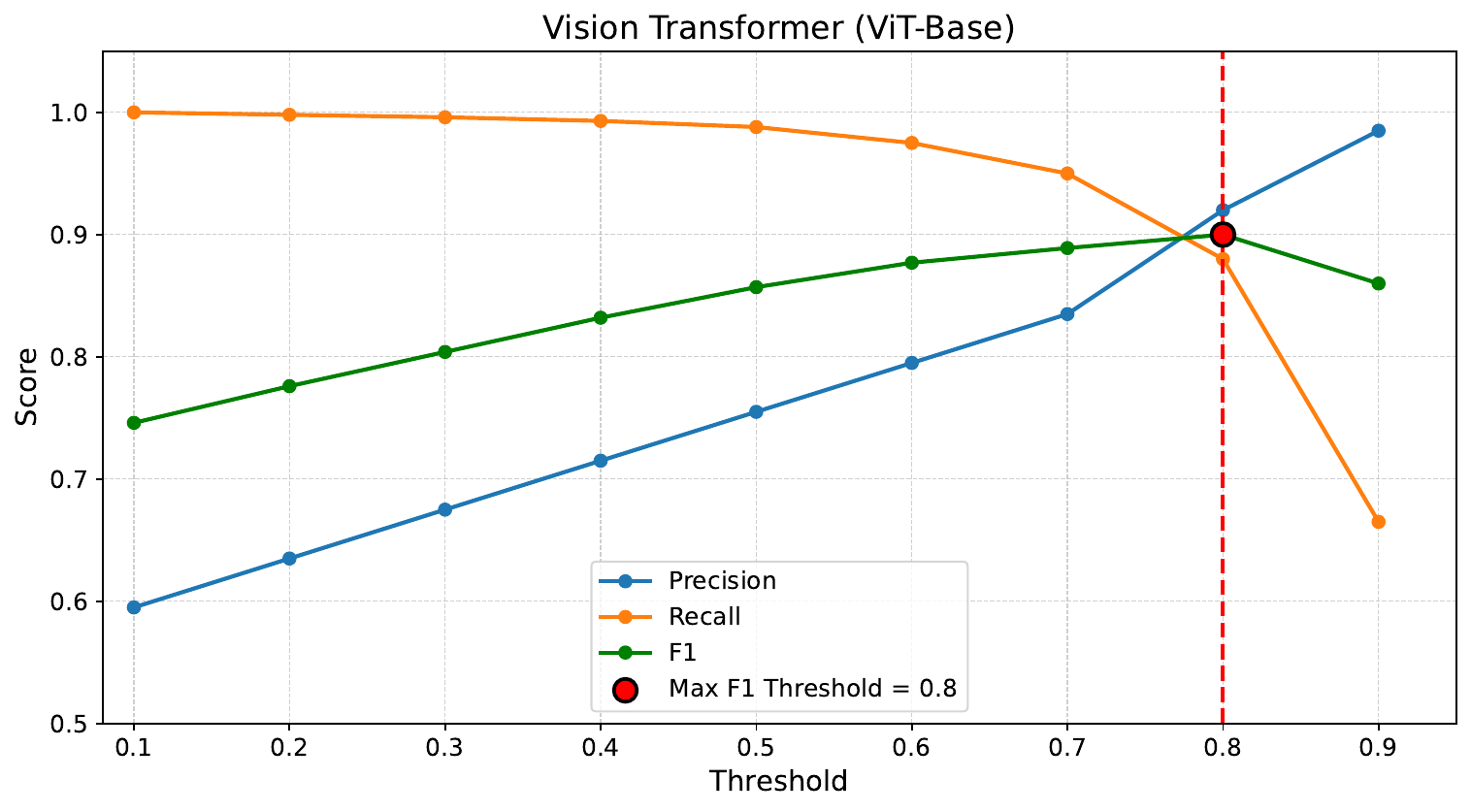}
\caption{Threshold analysis of ViT-Base.}
\label{fig:threshold-vit}
\end{figure}

Table~\ref{tab:vit_threshold} and Fig.~\ref{fig:threshold-vit} illustrate precision, recall, and F1-score against the various threshold values for the \textit{ViT-Base} model. From the results, it is observed that the ViT-Base performs well overall, but its best F1-score is lower compared to ConvNeXt-Tiny. This means the balance between recall and precision is not as strong at the thresholds that matter most for deployment. As a result, ConvNeXt-Tiny offers a more stable and reliable operating point for real phishing-detection scenarios.

\begin{table}[hpt]
\centering
\caption{Threshold-based performance of ViT-Base}
\label{tab:vit_threshold}
\begin{tabular}{c c c c}
\hline
\textbf{Threshold} & \textbf{Precision (\%)} & \textbf{Recall (\%)} & \textbf{F1-score (\%)} \\
\hline
0.1 & 0.595 & 1.000 & 0.746 \\
0.2 & 0.635 & 0.998 & 0.776 \\
0.3 & 0.675 & 0.996 & 0.804 \\
0.4 & 0.715 & 0.993 & 0.832 \\
0.5 & 0.755 & 0.988 & 0.857 \\
0.6 & 0.795 & 0.975 & 0.877 \\
0.7 & 0.835 & 0.950 & 0.889 \\
\textbf{0.8} & \textbf{0.920} & \textbf{0.880} & \textbf{0.900} \\
0.9 & 0.985 & 0.665 & 0.860 \\
\hline
\end{tabular}
\end{table}

Along with detection performance, we also looked at how efficient each model is to run, since this affects how well it can be deployed in real systems. ConvNeXt-Tiny needs less computation and runs with a lower inference cost because it has a smaller parameter size and uses a simpler convolutional design. ViT-Base benefits from its ability to capture global context through self-attention, but this also makes it heavier and more expensive to run. When comparing both models using F1-score as the main measure, ConvNeXt-Tiny offers a better mix of strong detection performance and lower computational load, which makes it more suitable for real-world phishing detection setups.

Table~\ref{tab:metrics} summarises the results (precision, recall, and F1-score) at the optimised threshold of 0.8. ConvNeXt-Tiny came out on top for precision, recall, and F1-score. This means it detects more phishing pages while keeping false positives lower. Overall, it provides a stronger balance between detection accuracy and false-alarm control compared to ViT-Base.

\setlength{\tabcolsep}{16pt}
\begin{table}[htp]
\centering
\caption{Performance at optimal threshold.}
\label{tab:metrics}
\scriptsize
\begin{tabular}{l        c        c}
\toprule
\textbf{Metric } & \textbf{ConvNeXt-Tiny} & \textbf{ViT-Base} \\
\midrule
Precision      & 0.997 & 0.920 \\
Recall         & 0.984 & 0.880 \\
F1-score       & 0.992 & 0.900 \\
\bottomrule
\end{tabular}
\end{table}

The results clearly show that ConvNeXt-Tiny performs more consistently at the optimal threshold of 0.8. The F1-score of 0.992 reflects a strong balance between precision and recall, meaning the model is able to detect phishing pages effectively while keeping false alarms very low. In practical terms, this is important because blocking legitimate websites can disrupt users, while missing phishing pages poses security risks. The reported precision (0.997) and recall (0.984) indicate that the model manages this balance well.

Although ViT-Base achieves reasonable performance, its lower recall (0.880) suggests that it misses a higher number of phishing instances compared to ConvNeXt-Tiny. This difference may be related to the architectural characteristics of transformer-based models, which rely more on global context and may require larger datasets to reach their full potential.

Another important observation is the stability of ConvNeXt-Tiny across different threshold values. The changes in precision and recall are gradual rather than abrupt, indicating that the model is not overly sensitive to small shifts in the decision boundary. This behaviour is desirable in deployment scenarios, where operating conditions may vary.

From a practical standpoint, threshold tuning plays a central role in how the system would behave once deployed. In environments where security is the highest priority, the threshold could be adjusted to favour recall and minimise missed phishing attempts. On the other hand, in user-facing systems where excessive blocking is problematic, the threshold could be set to favour precision. The threshold-aware analysis presented in this study allows such flexibility, which strengthens the real-world relevance of the framework.


The primary objective of this study was to develop a deployment-oriented framework for image-based phishing detection that goes beyond accuracy-based evaluation. The results presented directly support this objective. By analysing model performance across multiple decision thresholds, we demonstrated that ConvNeXt-Tiny not only achieves high classification performance but also maintains stability under varying operating conditions. This confirms that threshold-aware evaluation provides a more realistic assessment of model behaviour compared to single-metric reporting.

\section{Conclusions and Future Directions}\label{sec:conclusions}

This paper provided an end-to-end system of image phishing detection using screenshots of webpages. The system addressed the shortcomings associated with text and URL-based phishing detection on visually deceptive attacks. The two models, ConvNeXt-Tiny and ViT-Base, were compared under similar training conditions and using threshold-aware metrics. The two models demonstrated that ConvNeXt-Tiny was more accurate with the highest F1-score at the optimal threshold. Its strength is based on its convolutional architecture that captures local information such as logo-based, layout-based, and text/image-based features predominantly found on phishing websites. It consumes lower computational power, which makes it efficient to deploy. ViT-Base, on the other hand, benefits from global context modelling but showed more sensitivity to threshold changes and lower recall, which reduced its F1-score. This suggests that CNN-based models still work very well for visual phishing detection, especially in situations where both accuracy and efficiency matter.

We also plan to release the webpage screenshot dataset that was used in this research so that others can reproduce the results of the study and build on these results. Sharing the data with the public will allow for the development of benchmarks and research to advance in this domain. We also intend to look at alternative future research as well, which may merge CNNs and transformers or create additional phishing simulation datasets with new phishing examples and alternative web styles. These ideas could help enhance the robustness and overall performance of systems that depend on images in the detection of phishing attempts.
\bibliographystyle{IEEEtran}
\bibliography{Myref}

\vspace{12pt}

\end{document}